\documentclass[JBR]{jbr_author} 
\pdfoutput=1
\usepackage{graphicx}
\usepackage{amsmath}
\usepackage{dcolumn}
\usepackage{bm}
\usepackage[FIGTOPCAP]{subfigure}

\title{Evolution of central pattern generators\\ for the control of a five-link bipedal walking mechanism}

\articletype{Research Article}

\author{At{\i}l{\i}m~G\"{u}ne\c{s}~Baydin\inst{1}\email{atilim@alumni.chalmers.se}}

\institute{
     \inst{1} Complex Adaptive Systems,\\
	 Department of Applied Physics,\\
     Chalmers University of Technology,\\
	 SE-412 96 G\"{o}teborg, Sweden}

\abstract{Central pattern generators (CPGs), with a basis is neurophysiological studies, are a type of neural network for the generation of rhythmic motion. While CPGs are being increasingly used in robot control, most applications are hand-tuned for a specific task and it is acknowledged in the field that generic methods and design principles for creating individual networks for a given task are lacking. This study presents an approach where the connectivity and oscillatory parameters of a CPG network are determined by an evolutionary algorithm with fitness evaluations in a realistic simulation with accurate physics. We apply this technique to a five-link planar walking mechanism to demonstrate its feasibility and performance. In addition, to see whether results from simulation can be acceptably transferred to real robot hardware, the best evolved CPG network is also tested on a real mechanism. Our results also confirm that the biologically inspired CPG model is well suited for legged locomotion, since a diverse manifestation of networks have been observed to succeed in fitness simulations during evolution.}


\begin{document}
\maketitle


\section{Introduction}
The field of humanoid robotics has received an increasing interest over the last few decades \cite{Raibert1986,Adams2000}. In contrast to conventional robotic designs working in specialized environments (e.g. highly customized robotic arms in assembly lines), robots with anthropomorphic features are expected to be more adept in a growing number of environments in which they are to perform, as most of the real environments have been designed to suit human anatomy and needs \cite{Swinson2000}. Humanoid designs are also desirable from a human--robot interaction point of view: humans tend to interact and communicate better with human-like entities \cite{Yamaoka2007}.

Bipedal locomotion is a principal part of the research efforts in the field of humanoid robotics. The main motivation for studying bipedal locomotion, and walking robots in general, is that it is in many ways superior to conventional wheeled approaches on real terrain \cite{Huang2001} and in situations where robots need to accompany, or replace, humans. Another motivation for the research on bipedal walking robots is to gain a better understanding of the physiology of human locomotion \cite{Raibert1986}.

It has been suggested that bipedal walking mechanisms are more flexible in coping with obstacles in complex environments when compared with other walking mechanisms (quadruped, insectoid etc.) \cite{Collins2005}. But this comes with the cost of substantially reduced stability, which in turn asks for more sophisticated control approaches. While stability is the main incentive for designing better control methods, it is also of note that recent research has made progress with control methods focused on reducing the impact of falling in case the locomotion system fails \cite{Fujiwara2003,RuizDelSolar2009}. Techniques applied so far to bipedal walking include model-based control, such as inverted pendulum dynamics \cite{Sano1990}, the commonly used zero moment point (ZMP) control \cite{Kajita2003}, and biologically inspired control methods using neural networks \cite{Taga1991,Latham1992,Geng2006,Ijspeert2008}.

In this study we present the results of applying genetic algorithms (GA) for the evolution of \emph{central pattern generator} (CPG) type neural networks controlling bipedal locomotion. CPGs, with a firm basis in neurophysiologic experiments \cite{Rossignol2006,Ijspeert2008}, describe rhythmic motor patterns in terms of oscillatory outputs from units of mutually inhibiting neurons. CPG networks are mathematically modeled using a set of coupled differential equations with a given connectivity structure and set of parameters. In most studies involving CPGs, the networks are tailor-made for a specific application and the lack of generic methods and design principles for creating CPGs with desired behavior is acknowledged in the field \cite{Righetti2006,Ijspeert2008,Liu2008}. The main aim in this study is the application of GA optimization for the creation of CPG networks under fitness evaluations in a realistic physical simulation. We apply CPG control to a \emph{five-link planar bipedal walking mechanism} \cite{Taga1991,Lewis2005}, a minimalistic structure containing four actuators (corresponding to the hip and knee of each leg) and touching the ground with rounded extremities without ankles (Figure~\ref{FigureHardware2}). Presented mathematical model and simulation are based on the work by Taga et al. \cite{Taga1991}; and by using a minimal set of actuators and a planar support structure, we aim to restrict this current study to the smallest set of parameters and two dimensions, comparable to several existing studies of bipedal walking \cite{Latham1992,Pratt2001,Lewis2003}. We also present a test of the best evolved network on a physical five-link walking mechanism, to investigate whether CPG networks evolved using simulations can be acceptably transferred to real robot hardware.

After presenting background information on CPGs and the five-link walking mechanism in Section~\ref{SectionBackground}, the article continues in Section~\ref{ExperimentalSetup} with details of the experimental setup including the physical simulation and the CPG network. This is followed by a selection of obtained results in Section~\ref{ResultsAndDiscussion}, and the conclusions in Section~\ref{Conclusions}.

\section{Background}
\label{SectionBackground}

\subsection{Central Pattern Generators}
Neurophysiologic studies on animals suggest that their nervous systems incorporate specialized oscillatory neural circuits, termed central pattern generators (CPG) that are responsible for most of the rhythmic movements produced by the organism, including locomotion \cite{Delcomyn1980,Rossignol2006}. These neural circuits are capable of producing high-dimensional rhythmic output for motor control of muscle groups while receiving only simple low-dimensional input signals from the central nervous system modulating their activity \cite{IJspeert2008}. In addition to this reduction of dimensionality of control signals, another defining characteristic of CPGs is that they are capable of sustained rhythmic activity without any dependence on oscillations in their input. In experiments on living animals, it has been shown for numerous cases that these neural circuits produce sustained rhythmic activation patterns even when they are in isolation from external stimuli \cite{Marder2001}.

In the field of robotics, CPGs implemented as neural networks have been applied to the control of legged locomotion for bipedal, quadrupedal, and other designs. Studies about CPG controlled bipedal locomotion have often been inspired by the seminal work of Taga \cite{Taga1991}, and now have grown to involve a diverse range of subjects such as adaptive dynamic motion \cite{Komatsu2005}, sensory-motor coordination \cite{Heliot2008}, supervised learning of periodic behavior \cite{Righetti2006}, and nonlinear dynamics of CPG control \cite{Aoi2005}. CPGs have also been successfully applied to non-legged cases such as serpentine locomotion \cite{Conradt2003} and swimming \cite{Lachat2006}. It has also been demonstrated that CPG controllers can be implemented as analog electronic circuits \cite{Lewis2003}. Important advantages of legged robotic locomotion with CPGs are: being biologically inspired and suited for distributed implementation, having few control variables, and exhibiting stable limit cycle behavior resistant to perturbations, as compared to classical control approaches such as the widely used \emph{zero moment point} (ZMP) method, finite-state machines, or heuristic control \cite{Ijspeert2008}.

Mathematically, CPGs are modeled as systems of coupled differential equations, similar to models for other continuous-time artificial neural networks. A CPG network is composed of \emph{oscillatory units}, an arrangement of two mutually inhibiting neurons each becoming active in turn. An oscillatory unit has a natural frequency and amplitude of oscillation on its own (depending on oscillation parameters), but when several such units are interconnected, and in turn, bound to an external input, they tend to tune in to the frequency of the presented input. By connecting these oscillatory units in different ways, networks with complex frequency and phase relationships can be constructed, which are very suitable for the control of walking mechanisms. Several nonlinear oscillatory units have been used in CPG research, including the Hopf, Rayleigh, van der Pol, and Matsuoka oscillators \cite{Liu2008}. In this study, we use the \emph{half-center} oscillatory model by Matsuoka \cite{Matsuoka1985,Matsuoka1987} due to its simplicity, existing wide use in CPG robotics research \cite{Liu2008}, and especially its use for bipedal locomotion by Taga et al.\cite{Taga1991} that forms the inspiration for our approach. The details are given in Section~\ref{ExperimentalSetup}.

\subsection{Five-link Planar Bipedal Walking Mechanism}

It has been recently demonstrated that anthropomorphic mechanisms even without any actuation can stably walk down slopes in three-dimensions, by a controlled release of stored gravitational potential energy \cite{Collins2001}. These so-called \emph{passive dynamic walkers} suggested that bipedal gaits on level ground or upward slopes can be studied with a smaller number of actuators than previously considered. These observations fall under the newly developed concept of \emph{morphological computation} \cite{Pfeifer2009}, recognizing the role of mechanics in contributing to some aspects of the control processes involved in locomotion, or movement in general.

\begin{figure} 
\centering
\includegraphics[width=60mm]{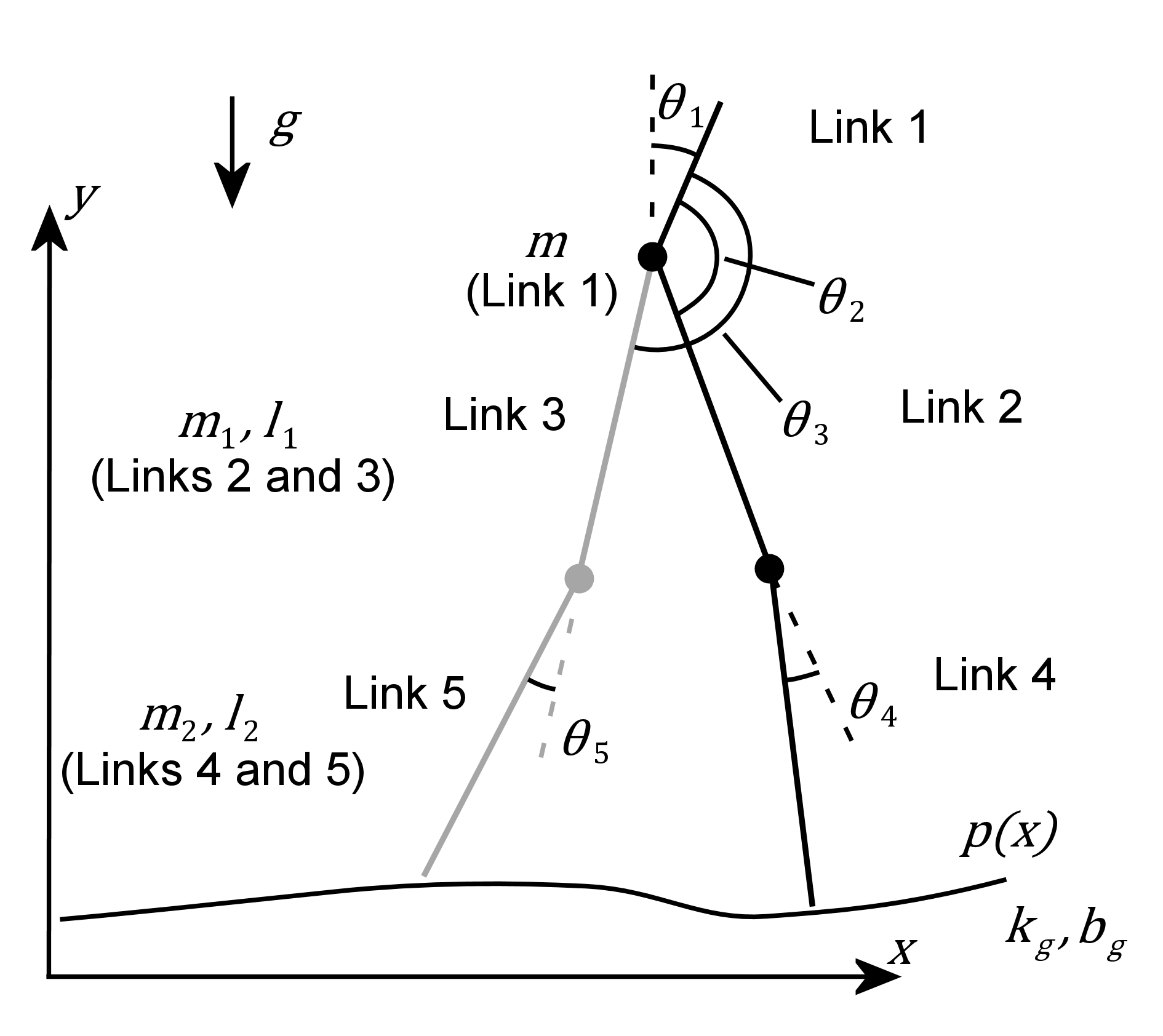}
\caption{The simulated five-link model. Black: right leg, gray: left leg. Link 1 is illustrated as an extrusion for ease of angular notation, but is in fact simulated as a point mass in the 2-dimensional plane, i.e. a linear connection between links 2 and 3 extending in the dimension perpendicular to the figure.}
\label{FigureModel}
\end{figure}

The five-link mechanism (Figure~\ref{FigureModel}), lacking feet and having only four actuators (two for the hips and two for the knees), is one of the simplistic approaches for studying bipedal locomotion. This mechanism is often planar, that is, restricted to run in two dimensions by means of an attached lateral boom \cite{Lewis2005,Chevallereau2003}, as is the case in this study. This has the advantage of making the physical simulation and mathematical analysis of the gait less complicated, providing insight on some of the central processes that occur in the lateral plane during bipedal locomotion. Mechanisms of this type have been used on many occasions to study bipedal walking \cite{Collins2001,Lewis2003,Lewis2005,Komatsu2005,Liu2008} and running \cite{Chevallereau2003}. The five-link mechanism that we use in this study as a physics simulation and constructed hardware is comparable with the one used by Lewis et al. \cite{Lewis2005}, and also by Geng, Porr, and W\"{o}rg\"{o}tter \cite{Geng2006} and Pratt et al. \cite{Pratt2001}, in terms of general structure and the arrangement of lateral support. The mechanism is described in detail in Section~\ref{ExperimentalSetup}.

\section{Experimental Setup}
\label{ExperimentalSetup}

Our study is based on a reimplementation of the mathematical model of the five-link walking mechanism and its coupled CPG controller introduced in the seminal work of Taga et al. \cite{Taga1991}. Using our implementation of this mathematical model, we set up a physics simulation enabling realistic fitness evaluations of evolving CPG networks that control the simulated walking mechanism. The walking mechanism itself that we use is comparable to those in several existing studies \cite{Latham1992,Pratt2001,Geng2006}, most notably by Lewis et al. \cite{Lewis2005}. While the study of Lewis et al. is also based on a mechanism with four actuators coupled to a CPG network similar to ours, it is concerned with the implementation and hand tuning of a particular CPG network (with a given connectivity structure) on a custom VLSI chip, whereas our study is concerned with the automated design of a CPG network (including the parameters and connectivity) through evolutionary computation.

In this section we present descriptions of the CPG network, walking mechanism, and physics simulation that we use, followed by the details of the evolutionary algorithm and hardware implementation.

\subsection{CPG Controller}
\begin{figure} 
\centering
\includegraphics[width=55mm]{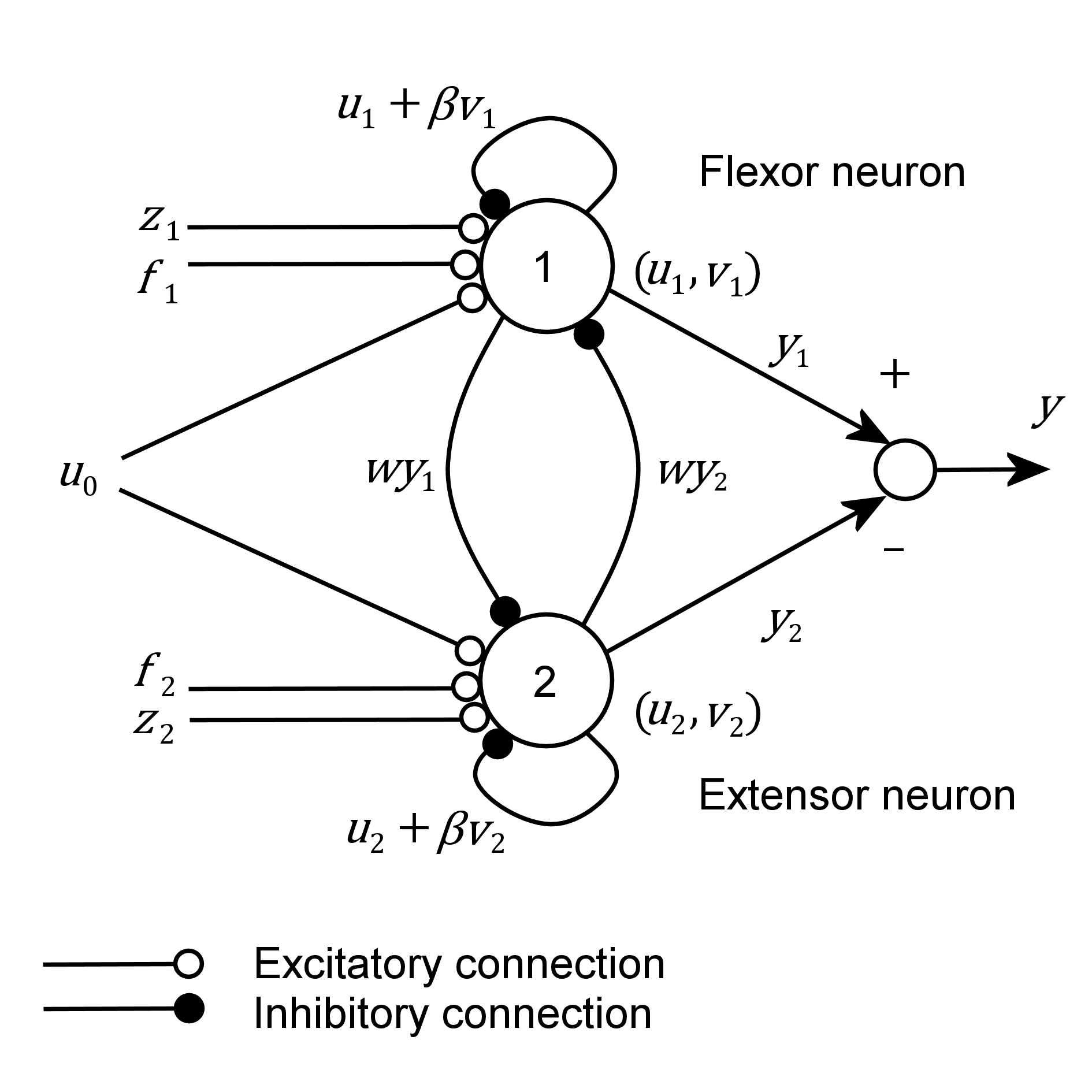}
\caption{Matsuoka's half-center model, as employed in the current study. Note that the inputs $f_{i}$ and $z_{i}$ to each neuron, both shown here to be excitatory, may either be excitatory or inhibitory.}
\label{FigureNeuron}
\end{figure}

The basis for the CPG controller used in this study is the half-center oscillatory unit known as Matsuoka's oscillator \cite{Matsuoka1985,Matsuoka1987}. The model is described by the following set of equations:

\begin{align}
\label{EquationOscillator}
\tau \dot{u}_1 &= u_0 - u_1 - w y_2 - \beta v_1 + z_1 + f_1 \nonumber\enspace{,}\\
\tau \dot{u}_2 &= u_0 - u_2 - w y_1 - \beta v_2 + z_2 + f_2 \nonumber\enspace{,}\\
\tau ' \dot{v}_1 &= - v_1 + y_1\enspace{,}\\
\tau ' \dot{v}_2 &= - v_2 + y_2 \nonumber\enspace{,}\\
y_i &= max(0, u_i), i = 1, 2 \nonumber\enspace{,}
\end{align}

where $u_i$ is the inner state, $v_i$ is the variable of self inhibition, $y_i$ is the output of the $i$th neuron, $u_0$ is a constant excitatory input to the oscillator, $\tau$ and $\tau '$ are time constants, $\beta$ is the coefficient of self inhibition, and $w$ is the weight of inhibitory connection between neuron 1 and neuron 2 (Figure~\ref{FigureNeuron}). The value of parameter $u_0$ has an effect on the oscillation amplitude, and the values of $\tau$ and $\tau '$ determine the natural oscillation frequency of the unit oscillator in the absence of an oscillatory input from other sources (which might be introduced by $z_i$ and $f_i$, as described further below).

One such oscillatory unit is responsible for the control of one mechanical joint in the walking mechanism, making a total of four unit oscillators in our setup with the five-link mechanism. Neuron 1 and neuron 2 of an oscillatory unit are respectively denoted \emph{flexor} and \emph{extensor} neurons for that joint, drawing on an analogy with the anatomy of muscular action in real joints. The output $y = y_1 - y_2$ of the oscillatory unit (Figure~\ref{FigureNeuron}) is used as the angular speed of the corresponding joint, after a linear scaling that is described in the following section.

$z_i$ in Eq.~(\ref{EquationOscillator}) represent the total input from other CPG unit oscillators in the controller network to the $i$th neuron of this unit oscillator, which might be excitatory (positive) or inhibitory (negative). This can be written as

\begin{equation}
z_i = \sum_j w_{ij} y_j\enspace{,}
\end{equation}

with $y_j$ representing the output of the $j$th neuron in the set of remaining unit oscillators in the network, and $w_{ij}$ is the connection weight existing in-between. 

Often, the input to the components of an oscillatory unit is arranged such that $z_1 = -z_2$, meaning that an internal network connection having the effect of, say, promoting the flexion movement of a joint (or equally, inhibiting the extension movement), should excite the flexor neuron and at the same time inhibit the extensor neuron of the corresponding oscillatory unit. The condition $z_1 = -z_2$ is not implicitly imposed in this study, and the GA implementation is free to determine the connection paths and types of connection to each of the neurons in a unit oscillator independently.

\begin{figure} 
\centering
\includegraphics[width=65mm]{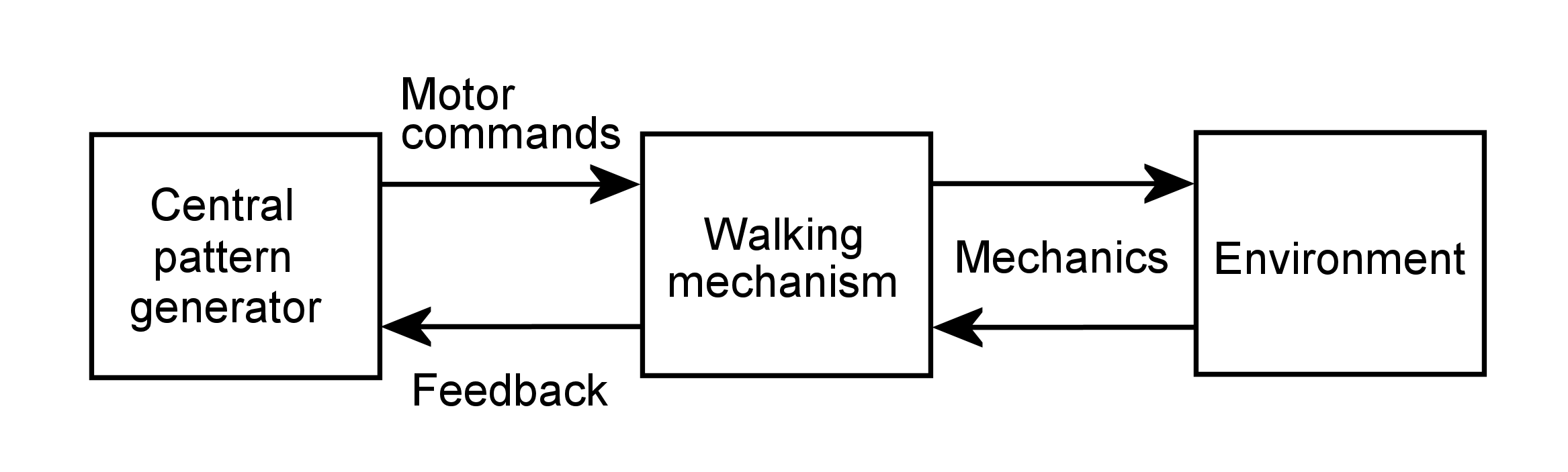}
\caption{An illustration of the concept of entrainment between the CPG network and the walking mechanism in the environment.}
\label{FigureEntrainment}
\end{figure}

$f_i$ in Eq.~(\ref{EquationOscillator}) represents the total feedback input to the $i$th neuron, in a similar fashion to $z_i$ described above. Feedback paths provide a means to maintain an adaptive mutual coordination, called \emph{entrainment}, between the CPG network and the walking mechanism subject to physics of the environment \cite{Taga1991}. This is achieved by a cyclic and continuous modification of oscillation characteristics and phase relations of the CPG network by the external inputs; and in turn, the commands sent by the CPG network moving the walking mechanism within the environment; and again, the effect of this on the CPG network through feedback (Figure~\ref{FigureEntrainment}). Feedback pathways between the mechanism and the CPG network are described in the following section.

\subsection{Walking Mechanism and the Physical Simulation}
\label{WalkingMechanismAndThePhysicalSimulation}

The five-link planar walking mechanism considered in this study consists of four actuators and the five links in-between: two actuators in the hip and two in the knee joints, two links for each leg, and the fifth link connecting the two legs. The mechanism has no supporting feet and no ankle joints, and touches the ground on two points at the end of its legs, which are rounded in the hardware implementation. The omission of feet is made possible by the increased stability of the two dimensional model: as long as the center of mass of the system lies between the two points of contact between the mechanism and the ground, any movement of the system ends in a stable state.

The mathematical model for physics simulation of the five-link mechanism is identical to that used by Taga et al. \cite{Taga1991}, with the simplification of omitting ankle actuators and joints, leaving four actuators in total. This is a design choice we make for seeking the simplest bipedal walking mechanism we can make subject to evolutionary optimization. Our implementation of this model essentially involves numerical integration of Newton--Euler equations describing the five rigid bodies constituting the mechanism in two dimensions and an impact model between the feet and the ground profile described as a spring--damper system, which is adapted from the model used by Raibert \cite{Raibert1984}.

Figure~\ref{FigureModel} presents an overview of the model. Angular relationships of the constituent links of the mechanism are all given with respect to Link 1, the link between the two hip joints standing at angle $\theta_1$ measured from the vertical. The thighs, Link 2 and Link 3, stand at $\theta_2$ and $\theta_3$ with respect to Link 1; and the shanks, Link 4 and Link 5, stand at $\theta_4$ and $\theta_5$ measured from the corresponding thighs they are attached to, respectively. The CPG controller network, described previously, is tied to the walking mechanism by the coupling of the output $y_i$ from the corresponding oscillatory unit of the CPG network (Figure~\ref{FigureNeuron}) as the angular speed $\dot{\theta}_i$ of each joint:

\begin{equation}
\dot{\theta}_i = \min(\max(y_i, -1), 1) \omega_{max}\enspace{,}
\end{equation}

where, before driving the joint, the output $y_i$ is scaled such that its absolute value is bounded by the maximum attainable rotational speed $\omega_{max}$ of the joints, dictated by the servo motor specifications of the hardware implementation (Section~\ref{Hardware}).

The mass of Link 1 connecting the hip joints is denoted $m$, and the link is simulated as a point mass on the simulation plane. $m_1$ and $l_1$ denote the mass and the length of thighs and $m_2$ and $l_2$ denote the mass and the length of the shanks for both legs. For both thighs and shanks, the center of mass is assumed to be halfway through their length. $k_g$ and $b_g$ represent the elasticity and damping coefficients of the ground impact model. In total, the implemented physics model has nine parameters describing the walking mechanism and its interaction with the environment, with the addition of gravitational acceleration $g$ and the maximum rotational speed $\omega_{max}$ of the joints.

The function $p(x)$ defines the ground profile $y = p(x)$, which provides a simple means to describe the environment the mechanism is simulated in, and makes it possible to define any form of terrain including slopes and simple obstacles. Even more complex cases can be introduced by defining $p(x)$ as a Fourier series with appropriate parameters.

The feedback pathways from the walking mechanism to the CPG controller network are based on six different features, which are the deviation from the vertical of each of the four joints and the state of the two feet. The feedback pathways are described by the following set of equations:

\begin{align}
\mbox{Left hip:}\nonumber\\
f_1 &= a_1(\theta_1 + \theta_3 - \pi) - a_1(\theta_1 + \theta_2 - \pi) + a_1 t_r\enspace{,}\\
f_2 &= -f_1\nonumber\enspace{,}\\
\mbox{Right hip:}\nonumber\\
f_1 &= a_1(\theta_1 + \theta_2 - \pi) - a_1(\theta_1 + \theta_3 - \pi) + a_1 t_l\enspace{,}\\
f_2 &= -f_1\nonumber\enspace{,}\\
\mbox{Left knee:}\nonumber\\
f_1 &= a_2 t_r (\theta_1 + \theta_2 + \theta_4 - \pi)\enspace{,}\\
f_2 &= -f_1\nonumber\enspace{,}\\
\mbox{Right knee:}\nonumber\\
f_1 &= a_2 t_l (\theta_1 + \theta_3 + \theta_5 - \pi)\enspace{,}\\
f_2 &= -f_1\nonumber\enspace{,}
\end{align}

where $t_l$ represents whether the left foot is on the ground ($t_l = 1$) or not ($t_l = 0$); and the same holds for $t_r$ corresponding to the right foot. The presence and strength of feedback to hips and knees are regulated by two coefficients, $a_1$ and $a_2$, and the values of these are included in the GA optimization process. This minimalistic approach has been observed to produce results comparable with former studies of this kind, involving greater numbers of coefficients (up to four for each feedback pathway) \cite{Taga1991}.

\subsection{Genetic Algorithms}
\label{GeneticAlgorithms}

For optimizing the parameters of the CPG controller network, including the internal connectivity structure and the presence and strength of feedback pathways from the walking mechanism, we employ a standard genetic algorithms (GA) implementation. Fitness evaluations of the population are done in the physics simulation of the five-link mechanism with a set of parameters not subject to evolution, arranged to match dimensions and masses of the physical parts forming the hardware implementation (Section~\ref{Hardware}).

We use a fitness function measuring the horizontal displacement of Link 1 (Figure~\ref{FigureModel}) between the beginning and the end of the evaluation. During fitness evaluations, we use a time integration step of $\Delta t = 10^{-5}$ s in the simulation. Each fitness evaluation lasts until either: (1) the gait becomes unstable\footnote{Gait stability can be proved by demonstrating stability of cycles in a phase space formed by $\theta_1 \dots \theta_4$ in Figure~\ref{FigureModel} by analyzing Poincar\'{e} sections of motion \cite{Aoi2005}. Here in GA evaluations we use a simpler definition by using ``stable'' to mean that the individual has not fallen to the ground (i.e. all of the five links in Figure~\ref{FigureModel} remain above ground profile $p(x)$) anytime during fitness evaluation.} and fails; or (2) the maximum length of time that we set as $10$ s (or $1000000$ time steps, given $\Delta t$) has passed. At the start of each fitness evaluation, the mechanism is released with a straight upright posture ($\theta_1 = \theta_4 = \theta_5 = 0$, $\theta_2 = \theta_3 = \pi$), from slightly above the ground.

\begin{figure} 
\centering
\includegraphics[width=60mm]{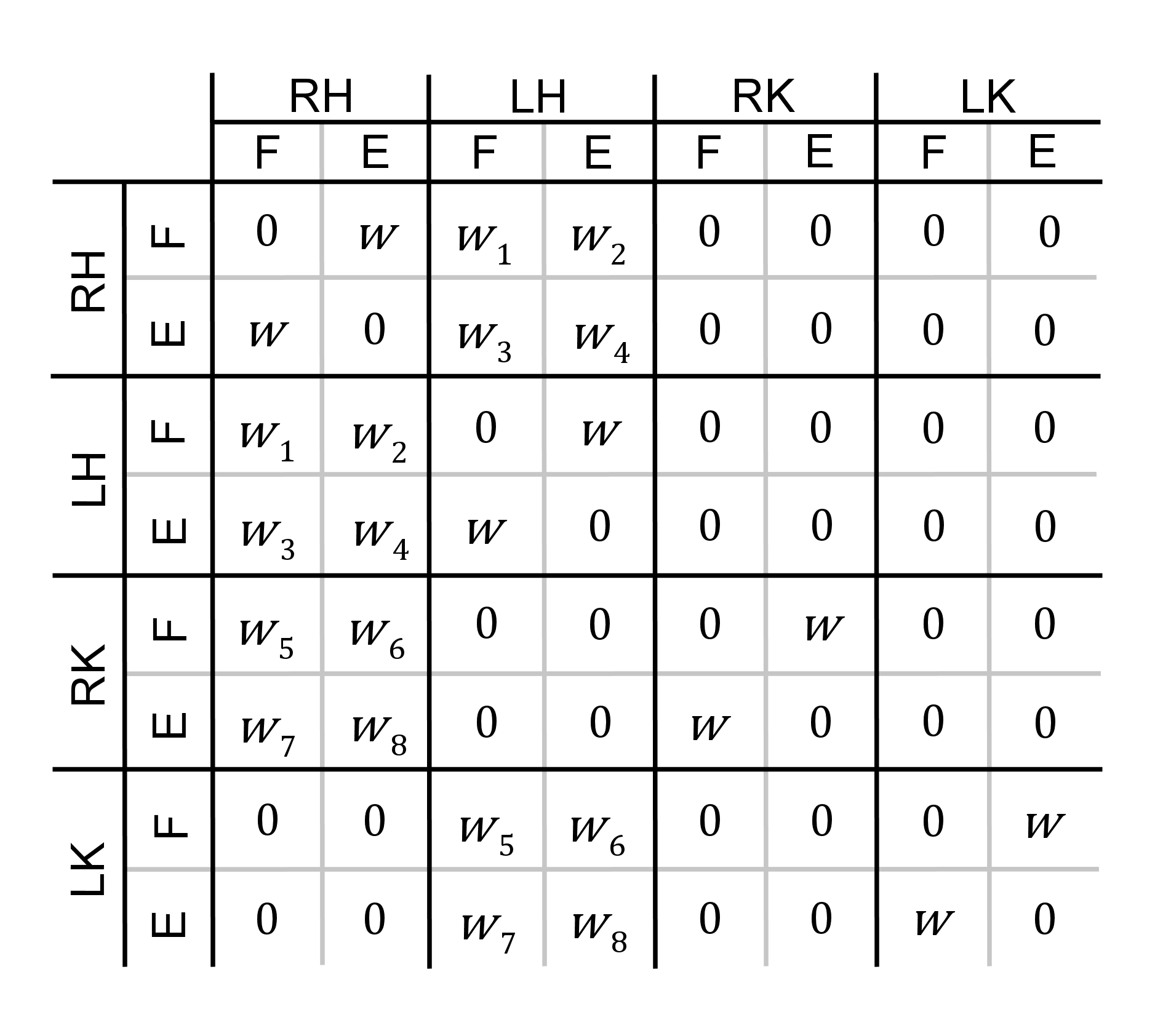}
\caption{Internal connectivity matrix of the CPG network. LH: left hip, RH: right hip, LK: left knee, RK: right knee, F: flexor neuron, E: extensor neuron. Knee neurons can be inhibited by the hip neurons on the same side, but the hips cannot be inhibited by the knees.}
\label{FigureConnectivity}
\end{figure}

We impose several restrictions on where connections might be present between the oscillatory units in the CPG network, with the aim of making use of the expected symmetry in the system (the bilateral symmetry of the walking mechanism and the expected behavioral symmetry of the human-like gait). We achieve this by the connectivity matrix presented in Figure~\ref{FigureConnectivity}. It shows, for a certain neuron (a row), the weights of incoming connections from other neurons (columns) in the network. This essentially reduces the network connectivity parameters to be optimized into the weight set $w_1$ to $w_8$.

In addition to these, for each evaluated individual, the encoded weights $w_1$ to $w_8$ are modified by a set of corresponding multipliers $w_1^*$ to $w_8^*$, which are filtered with the Heaviside step function to become either $0$ or $1$, providing a simple way for the evolution process to turn a given connection ``on'' or ``off''. That is

\begin{align}
w_i &\leftarrow w_i H(w_i^*)\enspace{,}\\
H(n) &= \begin{cases}
	0, & \text{if} n < 0\enspace{,}\\
	1, & \text{if} n \geq 0\enspace{.}
\end{cases}
\end{align}

The parameters $a_1$ and $a_2$ describing the feedback pathways are also coupled with parameters $a_1^*$ and $a_2^*$ in the same manner. We introduce these ``on/off'' multipliers to provide evolution with another layer of direct control over the network topology between the nodes, a GA technique adapted from the concept of non-expressed genetic code, i.e. intron, in genetics \cite{Wineberg1996}. The connection weight between coupled flexor and extensor neurons is given by $w$, which is the same for every oscillatory unit.

Note that the connections between the hip and knee unit oscillators of a side are unidirectional, as in Taga et al. \cite{Taga1991}: There might be an effect on knee unit oscillators from the corresponding hip unit oscillator on its side, but not the other way (correspondingly, the connectivity matrix in Figure~\ref{FigureConnectivity} is asymmetric).

In total, the encoding scheme describing individuals consists of 25 real numbers coding the parameters of the CPG network; the presence, strength, and nature (i.e. inhibitory or exhibitory) of internal connectivity between different oscillatory units; and the presence and strength of the feedback pathways described in the previous section. The encoding scheme includes: $w$, $u_0$, $\tau$, $\tau '$, and $\beta$ as the internal parameters of the unit oscillators; $w_1$ to $w_8$ and $w_1^*$ to $w_8^*$, the elements of the connectivity matrix and multipliers; and $a_1$, $a_2$, $a_1^*$, and $a_2^*$, the feedback coefficients and multipliers. The values of $\tau$ and $\tau '$ are directly used in the hip unit oscillators, whereas in the knee unit oscillators, these values are taken as $\tau / 2$ and $\tau ' / 2$, resulting in an oscillation frequency twice that of the hip joints, as is the case in a real human's gait \cite{Taga1991}. We use random initialization of all encoded variables in the initial population.

\subsection{Hardware}
\label{Hardware}

\begin{figure} 
\centering
\includegraphics[width=60mm]{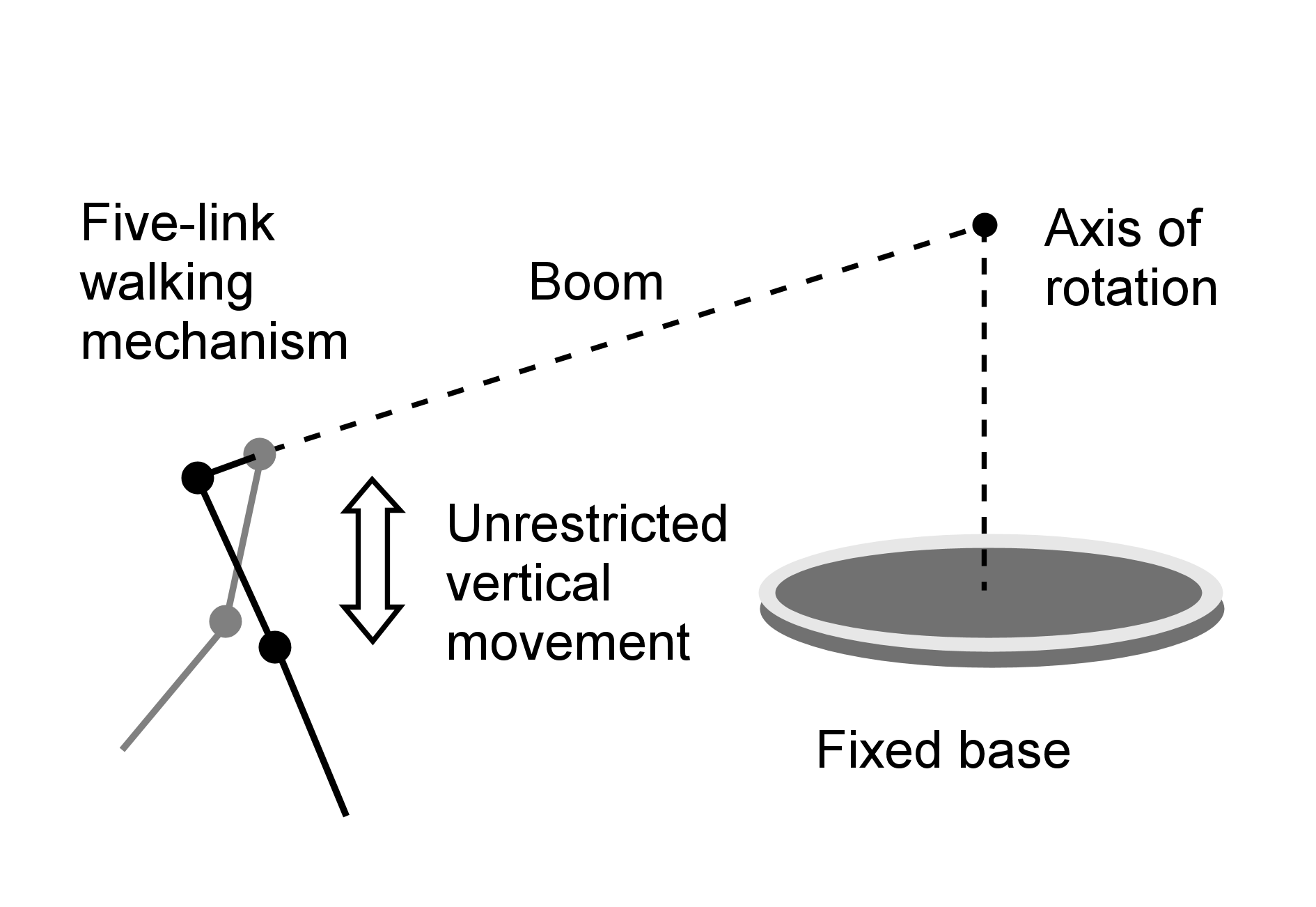}
\caption{The five-link planar walking mechanism and the accompanying support structure.}
\label{FigureHardware}
\end{figure}

With the intention of conducting a preliminary test about whether results we have from evolution in simulations can be acceptably transferred to real robot hardware, we construct a hardware implementation of the five-link mechanism. As with the physical simulation described previously, this mechanism needs to be run in two dimensions, and this is achieved, as in many previous studies \cite{Pratt1998,Lewis2005,Chevallereau2003,Geng2006}, by an attached lateral boom freely rotating around a pivot (Figure~\ref{FigureHardware}). The boom, of approximately $1.5~\textnormal{m}$ of hollow and lightweight plastic, restricts the movement of the mechanism to a spherical surface approximating motion in two dimensions, given the boom radius is sufficiently large compared to the mechanism. This support has minimal effect on the vertical stability of the mechanism in the \emph{sagittal plane} (the plane perpendicular to hip joint axes, dividing the body into left and right halves) and the dynamics are assumed to approximate the two-dimensional simulation.

\begin{figure} 
\centering
\includegraphics[width=60mm]{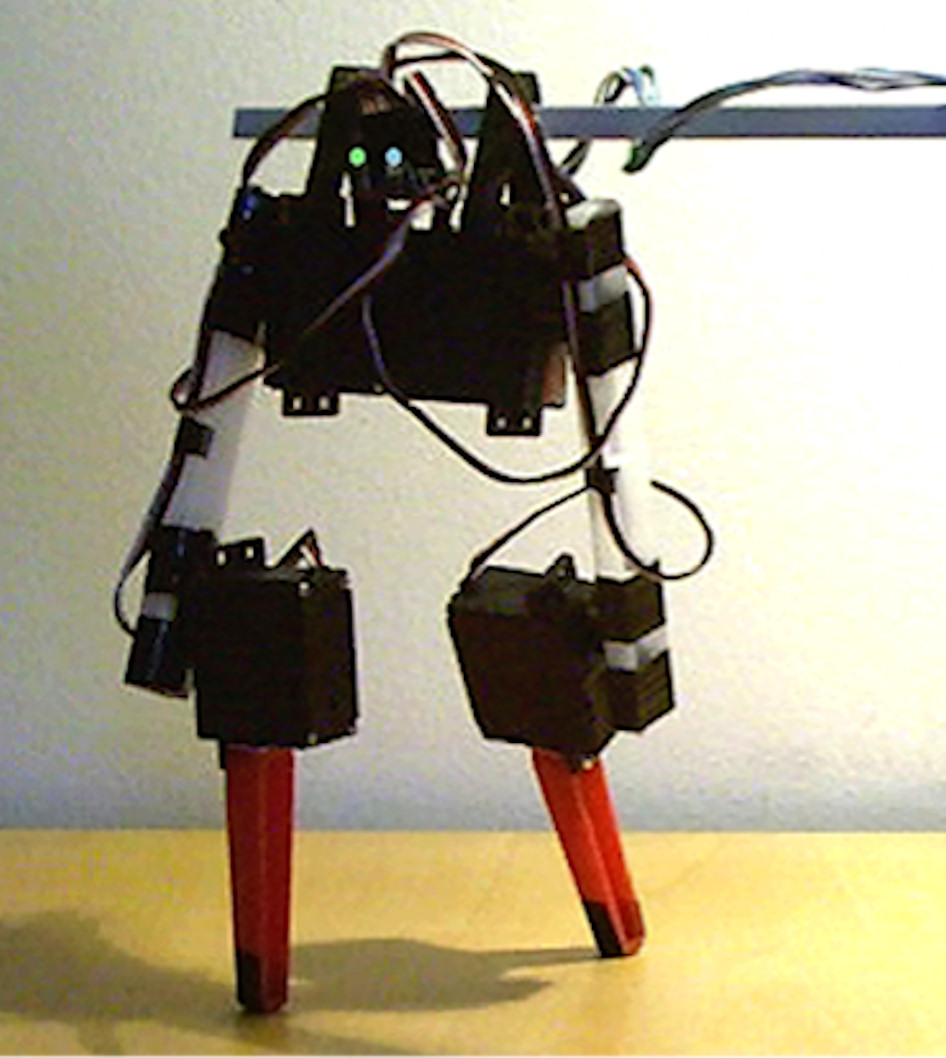}
\caption{The constructed five-link walking hardware. The mechanism is about $20~\textnormal{cm}$ ($7.87~\textnormal{inches}$) high and weighs slightly more than $200~\textnormal{grams}$ ($0.44~\textnormal{pounds}$}
\label{FigureHardware2}
\end{figure}

The body of the robot is constructed out of hard plastic parts as found fit for the purpose. Four standard servo motors are used as actuators (rated with $\omega_{max} = 5.51~\textnormal{radian}/\textnormal{s}$, see Section~\ref{WalkingMechanismAndThePhysicalSimulation}), with the motors in hip joints having a range of movement of $180^\circ$, and the motors in the knee joints $90^\circ$ (Figure~\ref{FigureModel}). The rounded tips of the legs making contact with the ground are covered with rubber tape for increasing the grip and avoiding slippage. The resulting mechanism, including the boom, is comparable to the one used by Geng et al. \cite{Geng2006}, and to the commercially available five-link robot ``Red-Bot'' used by Lewis et al. \cite{Lewis2005} and other studies. The finished mechanism (Figure~\ref{FigureHardware2}) is about $20~\textnormal{cm}$ ($7.87~\textnormal{inches}$) high and weighs slightly more than $200~\textnormal{grams}$ ($0.44~\textnormal{pounds}$).

The hardware is driven by a dedicated interface circuit attached to the host computer (running the CPG controller and GA evaluations), based on the Parallax Basic Stamp 2\texttrademark microcontroller. It communicates with the host computer using RS232 serial communications protocol.

\section{Results and Discussion}
\label{ResultsAndDiscussion}

\subsection{Simulation and GA}

At first, we tried a hand-tuning approach to see if a stable gait walking on level ground can be produced by manual adjustment of CPG parameters, using previous results by Taga et al. \cite{Taga1991} as the starting point. This proved to be very hard to achieve, and in the very rare cases where the mechanism could be made to walk, the gait seemed unnatural, and eventually destabilized into a fall after just three or four steps in all cases.

\begin{figure}
	\centering
	\subfigure[\hspace{5cm}]{
		\includegraphics[width=60mm]{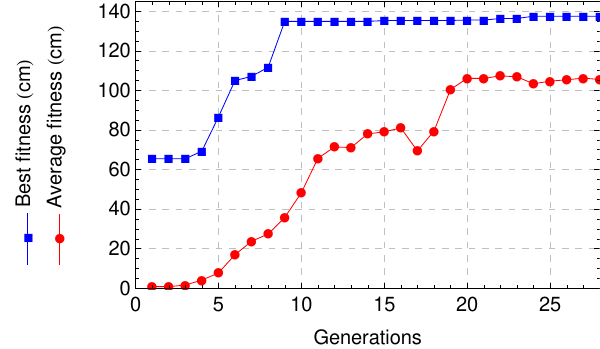}
		\label{FigureEvaluationA}
	}
	\subfigure[\hspace{5cm}]{
		\includegraphics[width=60mm]{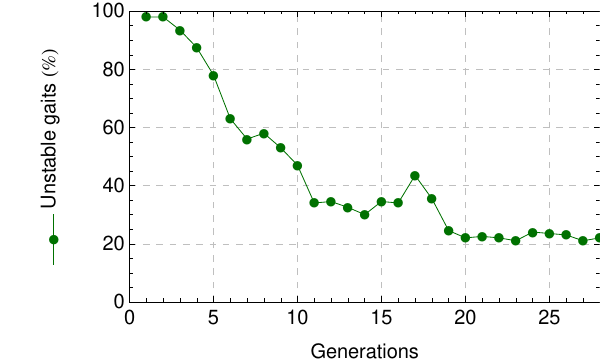}
		\label{FigureEvaluationB}
	}
	\caption{Plots of (a) the best and average fitnesses obtained by individual CPG networks and (b) percentage of unstable gaits, as evolution progresses.}
	\label{FigureEvaluation}
\end{figure}

Moving on to GA experiments, with the parameter set summarized in Table~\ref{TableParameters}, we observe stable gaits after just a few generations of fitness evaluations. After approximately 10 generations, the best fitness asymptotically reaches a plateau that should be determined by limiting factors such as the maximum speed of the actuators and the mass and dimensions of the mechanism, after which improvements are minor. Figure~\ref{FigureEvaluation}(a) gives a plot of the best and average fitnesses during a typical GA run, with the simplest fitness measure promoting the distance moved\footnote{During the course of our experiments, we have also introduced several other objectives to the fitness measure, such as promoting an upright posture and putting an upper boundary to the height of the mechanism from the ground, to prevent jumping. But, in the end, the improvement provided by these were minimal in all cases, and each additional fitness objective introduced additional exceptions to deal with.}. 

\begin{table}
\renewcommand{\arraystretch}{1.4}
\centering
\caption{\label{TableParameters}
GA parameters used in the evolution of CPG networks.}
\small
\begin{tabular}{m{20mm} m{60mm}}
\hline
Parameter & Value / description \\
\hline
Population size & $200^\textnormal{1}$ \\
Chromosome encoding & $25$ real numbers directly encoding CPG parameters \\
Selection & Tournament, with size $8$, probability $0.75$ \\
Crossover & Two points, with probability $0.8$ \\
Mutation & Simple one-point, with probability $0.3$ \\
\hline
\multicolumn{2}{p{80mm}}{$^\textnormal{1}$Made up of individuals reproduced from previous generation and individuals created by crossover (determined by the crossover probability). Elitism was used (the best individual was always kept).}
\end{tabular}
\end{table}

\begin{figure*}
\centering
\includegraphics[width=150mm]{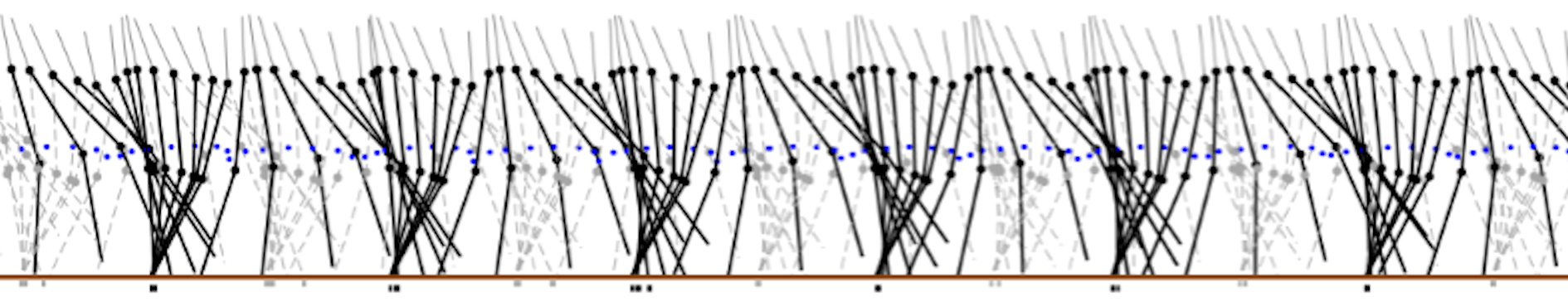}
\caption{The best evolved gait. Black: right leg, dashed gray: left leg. The dots near the center indicate the trajectory of the center of mass. The orientation of Link 1, hence the whole structure, is represented by the short gray line extruding upwards from the hip. The marks below the ground line represent contacts by the left foot (gray) and right foot (black). Direction of movement is to the right.}
\label{FigureBestGait}
\end{figure*}

\begin{figure*}
\centering
\includegraphics[width=120mm]{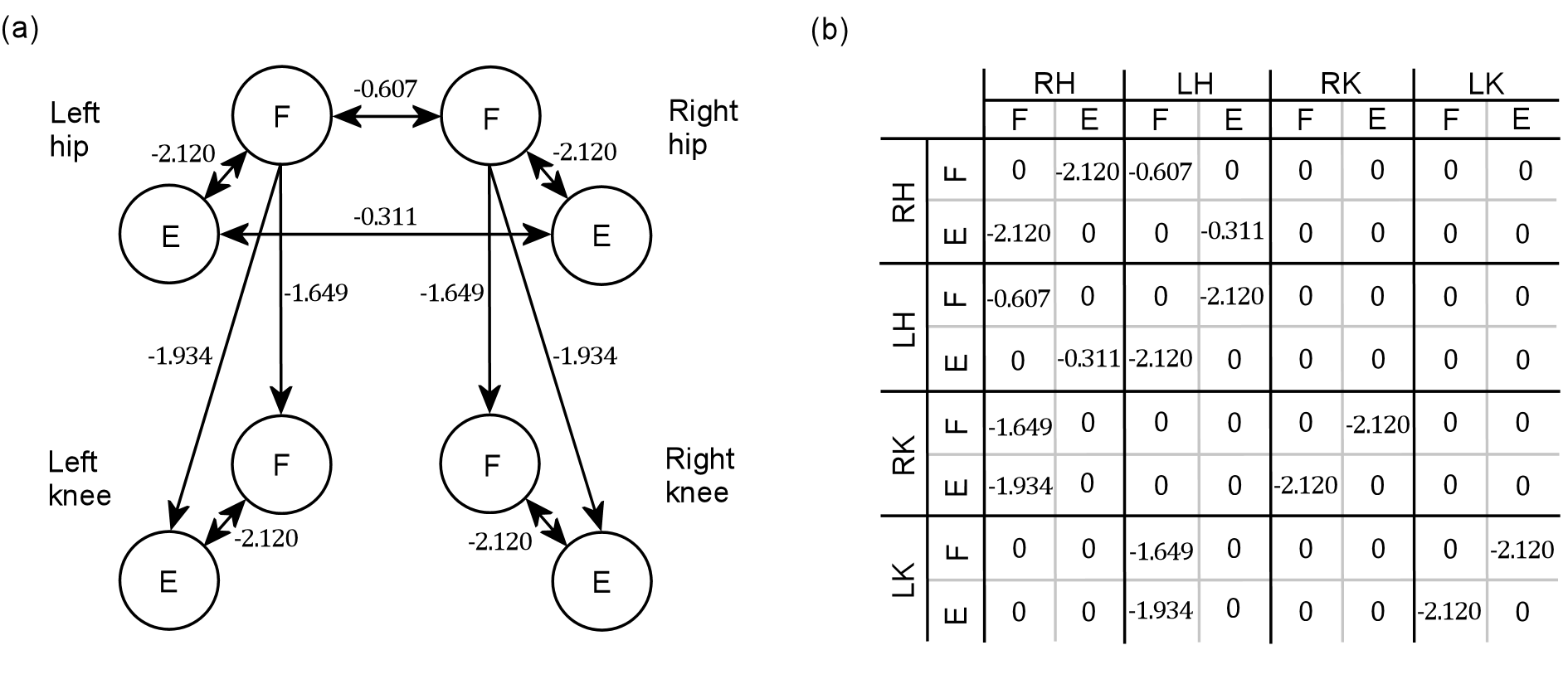}
\caption{(a) The best evolved CPG network structure. Arrowheads denote direction of connections, which are all inhibitory. (b) The connectivity matrix of the best evolved CPG network. F: flexor neuron, E: extensor neuron, LH: left hip, RH: right hip, LK: left knee, RK: right knee.}
\label{FigureBestNetwork}
\end{figure*}

The best gait evolved on level ground is represented in Figure~\ref{FigureBestGait}. The CPG parameters resulting in this gait are given in Table~\ref{TableBestNetwork} and the CPG network structure of the individual is given in Figure~\ref{FigureBestNetwork}. The resulting network uses six connections out of the possible sixteen. The snapshots in Figure~\ref{FigureBestGait} are $5.55 \times 10^{-2}$ s ($55$ ms) apart (cf. the time resolution of the simulation $10^{-5}$ s, or $0.01$ ms) and each step (either left or right) takes approximately $0.83$ s. The horizontal distance attained by this gait is $137.44$ cm in $10$ s, giving a speed of approximately $0.13$ m/s. Scaling this up to average human dimensions (by a factor of $168 / 20$, from the $20$ cm height of the mechanism and an average human height of $168$ cm \cite{Ellis2001}) gives a speed of $1.09$ m/s, which is significantly close to the typical human walking speed of approximately between $1.2$ m/s and $1.5$ m/s \cite{Carey2005}. The multiple foot contacts during the gait indicate a degree of bouncing occurring due to the spring--damper model of the ground that we employ. We also note that: (1) all cycles shown in the figure are not exact replicas of previous ones (evidenced by the ground contact points in each cycle) and (2) still, in the long run, the gait is sustained. This provides evidence for the concept of entrainment that we discussed in Section~\ref{ExperimentalSetup}: the variations in the system are kept under control by the continuing mechanical interaction of the mechanism and the environment, in contrast with an exact execution of a set trajectory in heuristic control approaches. This gait has also been tested in simulations with slightly inclined or declined (up to $5^\circ$) ground profiles $p(x)$ with success. 

\begin{table}
\renewcommand{\arraystretch}{1.4}
\centering
\caption{\label{TableBestNetwork}
Parameter set of the CPG network producing the best resulting gait. The values of $w_1, \dots, w_8$, $a_1$, and $a_2$ are after threshold operation with $w_1^*, \dots, w_8^*$, $a_1^*$, and $a_2^*$.}
\small
\begin{tabular}{m{20mm} m{60mm}}
\hline
Parameter & Value / description \\
\hline
$\tau$ & $0.285$ (hips), $0.143$ (knees) \\
$\tau '$ & $0.302$ (hips), $0.151$ (knees) \\
$w$ & $-2.120$ \\
$\beta$ & $3.078$ \\
$u_0$ & $0.805$ \\
$w_1, \dots, w_8$ & $-0.607, 0, 0, -0.311, -1.649, 0, -1.934, 0$ \\
$a_1, a_2$ & $0.124, 0.770$ \\
\hline
\end{tabular}
\end{table}

The overall result is that the CPG approach to human-like bipedal walking shows great versatility. Many different gaits were observed during the course of GA fitness evaluations, and it was particularly striking that even in the first generation (where the CPG parameters and connections are completely random by definition) there were individuals with stable-looking gaits, even if these were generally not sustained until the maximum allowed evaluation time of $10$ s. A video compilation of several fitness evaluations with varying degrees of success is provided online\footnote{The video file is available at \url{http://arxiv.org/src/0801.0830v8/anc/BaydinCPGEvolution2011.avi}}, also demonstrating the progress of evolution with passing generations.

The fraction of unstable gaits in the population drops sharply during approximately the first 10 generations and stabilizes around $20\%$ (Figure~\ref{FigureEvaluation}(b)). This level is slightly lower than the mutation rate of $30\%$ responsible for introducing randomness into the system, again suggesting that a portion of randomness does not result in total gait failure. This is suggestive of an inherent aptness of the CPG network for the task of walking. Virtually in all runs (with level and slightly inclined or declined (up to $5^\circ$) ground profiles), a stable gait was eventually established after five or six generations of the GA evaluation.

Also, occasionally during the course of GA evaluations, there emerge successful individuals with parameters $a_1$ and $a_2$ close to zero. This effectively means that these individuals are not making use of the feedback pathways, yet still are able to exhibit a stable gait. The existence of such individuals suggests that---at least on level terrain and without any obstructions---CPG control is capable of producing stable gaits with or without input from the environment. This is in agreement with neurophysiological studies of \emph{fictive locomotion}, providing evidence that rhythms can be centrally generated without requiring sensory information \cite{Ijspeert2008}. During the GA run that produced Figure~\ref{FigureEvaluation}, there were on average 4.73 individuals (out of 200) per generation with both $\{a_1^* = 0$ or $\left|a_1\right| \le 10^{-2}\}$ and $\{a_2^* = 0$ or $\left|a_2\right| \le 10^{-2}\}$ that exhibited a stable gait.

In addition, for the networks utilizing feedback, another result that we observe frequently is that knee oscillatory units make use of feedback more than the hip units do. The best CPG network presented in Table~\ref{TableBestNetwork} is a good example of this, where feedback strength to the knees $a_2$ is significantly larger than that to the hips $a_1$. This is indicative of a complex relationship between the controller network, the mechanism, and the environment. Because the neural connections between the hip and knee oscillators are unidirectional (i.e. no signal from the knees reach the hips), our explanation is that feedback received by the knees is propagated up to the hips through mechanical means. This provides strong evidence for the concept of \emph{morphological computation}, which states that materials forming the body and their action under the physics of the environment can take over some of the processes that had been conventionally attributed to control \cite{Pfeifer2009}.

\subsection{Hardware}

To observe whether results evolved under fitness evaluations in simulation can be acceptably transferred to real robot hardware, we conducted preliminary tests of the evolved CPG networks on the real mechanism. The mechanism was run on level ground, and correspondingly, results evolved for level ground were tried on the hardware. For establishing stability of the gaits on the hardware, we had to apply a linear scaling between the output of all unit oscillators running on the computer and the input to the microcontroller sending commands to the servo motors. Our interpretation of why this was needed is that it compensates for the difference in the speed response of the joints in simulation and the real servo motors. This scaling is the same for all joints, and therefore does not destroy the phase relations in the CPG output forming the gait evolved in the simulation. After this transformation, the runs generally have reproduced, to an acceptable degree, the walking speed attained by the best evolved gait in simulation, walking with approximately $0.2$ m/s speed (cf. $0.13$ m/s in simulation).

One particular drawback in the hardware runs is introduced by the laterally attached support structure. Even if there is an expected degree of asymmetry caused by the slightly unequal distances covered by the left and right feet (Figure~\ref{FigureHardware}), the impact of this on the gait balance was larger than what was anticipated. While this can be averted by using a sufficiently long boom compared to the distance between the legs (with the disadvantage of introducing more weight onto Link 1, prompting a change in parameter $m$ in Figure~\ref{FigureModel}), another solution, such as a linear overhead structure with a pin-in-slot mechanism used by Komatsu and Usui \cite{Komatsu2005} for a similar five-link mechanism, might prove to be a better option. This problem can also be addressed with a \emph{load compensation} \cite{Peterka2004} approach, which is not currently taken into account in our experimental design, that would dynamically apply corrective torques on the joints.

\section{Conclusions and Future Work}
\label{Conclusions}

We presented an approach where the connectivity and oscillatory parameters of a CPG network was subject to GA optimization, with fitness evaluations in a realistic simulation with accurate physics. We applied this technique to a five-link walking mechanism to demonstrate its feasibility and performance. To observe whether evolved networks can be acceptably transferred to real robot hardware, we also presented preliminary tests on a real mechanism. Our results also confirm that the biologically inspired CPG model is well suited for legged locomotion, since a diverse manifestation of networks have been observed to succeed in fitness simulations during evolution. We chose the five-link bipedal mechanism as a simple model for demonstrating our approach, but the study is easily extensible to the design of CPG networks controlling more complicated mechanisms.

During GA fitness evaluations, we observed a variety of individuals lacking feedback pathways, but still able to walk. This result indicates that the CPG approach, together with the structural constraints we imposed with the connectivity matrix (Figure~\ref{FigureConnectivity}), is inherently able to sustain a stable gait without any input from the environment; and if an even more simplistic model of CPG control were needed, it would be feasible to leave out the feedback terms from Eq.~\ref{EquationOscillator}. But one should note that a lack of feedback would be detrimental in any environment beyond the perfectly regular and planar one we simulated for fitness evaluation, as there is strong experimental evidence that feedback contributes to stability and load compensation \cite{Prochazka1997} and is needed to adapt to irregularities in the environment, such as varying ground slope or obstacles.

For future work, as a straightforward addition to this study, we plan to analyze the performance of our approach under: (1) the presence of obstacles; and (2) in dynamically changing environments. Regarding the evolutionary optimization of CPG controllers, our approach can be combined with novel CPG control approaches. The most notable among those we consider is the \emph{predictive and reactive tuning} technique by Prochazka and Yakovenko \cite{Prochazka2007} that introduces a regulation of CPG phases by rules combining sensory input, which, we envision, can be represented and optimized as individual genetic programming (GP) trees.

Regarding the physical characteristics of the walking hardware in this study, we consider the possibility of modifying the five link mechanism to walk in three dimensions without any support structure, by using wide feet ensuring lateral stability as used by Collins \cite{Collins2001}. It would be also interesting to approach the bipedal walking problem from an evolutionary robotics perspective \cite{Bongard2003}, by including the physical parameters of the mechanism (e.g. the sizes and masses of the links) in the evolutionary process to optimize bipedal walking hardware designs in environments with different characteristics.

\begin{acknowledgments}
For his support in this work, we wish to express gratitude to Dr. Krister Wolff at the Department of Applied Mechanics, Chalmers University of Technology, G\"{o}teborg, Sweden.
\end{acknowledgments}

\bibliographystyle{unsrturl}
\bibliography{BaydinHRCPG}

\end{document}